\documentclass[twocolumn, switch]{article} % Method A for two-column formatting

\usepackage{preprint}
\pdfoutput=1
\usepackage{amsmath, amsthm, amssymb, amsfonts}

\usepackage[numbers,square]{natbib}
\usepackage[utf8]{inputenc}	% allow utf-8 input
\usepackage[T1]{fontenc}	% use 8-bit T1 fonts
\usepackage{booktabs} 		% professional-quality tables
\usepackage{nicefrac}		% compact symbols for 1/2, etc.
\usepackage{microtype}		% microtypography
\usepackage{lineno}		% Line numbers
\usepackage{float}			% Allows for figures within multicol
\usepackage{multicol}		% Multiple columns (Method B)

\usepackage{lipsum}		%  Filler text

\usepackage{titlesec}
\titlespacing\section{0pt}{12pt plus 3pt minus 3pt}{1pt plus 1pt minus 1pt}
\titlespacing\subsection{0pt}{10pt plus 3pt minus 3pt}{1pt plus 1pt minus 1pt}
\titlespacing\subsubsection{0pt}{8pt plus 3pt minus 3pt}{1pt plus 1pt minus 1pt}

\usepackage{times}
\usepackage{helvet}
\usepackage{courier}
\usepackage{amsmath}
\usepackage{amssymb}
\usepackage{graphicx}
\usepackage{booktabs}
\usepackage{adjustbox}
\usepackage{multirow} % 用于合并单元格
\usepackage{array}    % 用于列格式调整
\usepackage{algorithm}
\usepackage{algpseudocode}
\usepackage{subcaption}

\title{MemoryKT: An Integrative Memory-and-Forgetting Method for Knowledge Tracing}

\usepackage{eso-pic}
\usepackage{tikz}
\usepackage{xcolor}
\PassOptionsToPackage{colorlinks=true,linkcolor=gray,urlcolor=gray}{hyperref}

\newcommand{\AddMyWatermarks}{%
  \begin{tikzpicture}[remember picture, overlay]
    \node[rotate=90, color=gray!60, scale=1] at ([xshift=-4.05in,yshift=0in]current page.center) {%
      \href{https://doi.org/}{Publication doi}%
    };
    \node[rotate=90, color=gray!60, scale=1] at ([xshift=3.9in,yshift=0in]current page.center) {%
      \href{https://doi.org/}{Preprint doi}%
    };
    \node[color=gray!90, scale=1] at ([xshift=0in,yshift=-5in]current page.center) {%
      This is the author's accepted manuscript. The final version will appear in XXXX 2025, © XXXX 2025.%
    };
  \end{tikzpicture}%
}

\AddToShipoutPictureBG*{\AddMyWatermarks}

\usepackage{titling}
\usepackage{orcidlink}
\usepackage{footmisc}
\newcommand{\Author}[3]{% Name, ORCID, Institution
  \textbf{#1}\textsuperscript{#2},\ \orcidlink{#3} %
}

\author{
  \Author{Mingrong Lin}{1}{}\and
  \Author{Ke Deng}{2}{0000-0002-1008-2498}\and
  \Author{Zhengyang Wu}{1}{0000-0002-3171-4618}\and
  \Author{Zetao Zheng}{3}{0000-0002-7801-0378} \and
  \Author{Jie Li}{1}{}
}

\date{%
  \textsuperscript{1}School of Computer Science, South China Normal University, Guangzhou, China\\
  \textsuperscript{2}School of Computing, National University of Singapore, Singapore\\
  \textsuperscript{3}Sichuan Artificial Intelligence Research Institute, Sichuan, China\\[1em]
  \footnotesize \textbf{Corresponding author:} Zhengyang Wu\texttt{}\\
}

\begin{document}

\twocolumn[ % Method A for two-column formatting
  \begin{@twocolumnfalse} % Method A for two-column formatting

\maketitle
\thispagestyle{empty}

\begin{abstract}
Knowledge Tracing (KT) is committed to capturing students' knowledge mastery from their historical interactions. Simulating students' memory states is a promising approach to enhance both the performance and interpretability of knowledge tracing models. Memory consists of three fundamental processes: encoding, storage, and retrieval. Although forgetting primarily manifests during the storage stage, most existing studies rely on a single, undifferentiated forgetting mechanism, overlooking other memory processes as well as personalized forgetting patterns. To address this, this paper proposes memoryKT, a knowledge tracing model based on a novel temporal variational autoencoder. The model simulates memory dynamics through a three-stage process:
(i) Learning the distribution of students' knowledge memory features,
(ii) Reconstructing their exercise feedback, while
(iii) Embedding a personalized forgetting module within the temporal workflow to dynamically modulate memory storage strength.
This jointly models the complete encoding-storage-retrieval cycle, significantly enhancing the model's perception capability for individual differences. Extensive experiments on four public datasets demonstrate that our proposed approach significantly outperforms state-of-the-art baselines.
\end{abstract}
%\keywords{First keyword \and Second keyword \and More} % (optional)
\vspace{0.35cm}

  \end{@twocolumnfalse} % Method A for two-column formatting
] % Method A for two-column formatting

%\begin{multicols}{2} % Method B for two-column formatting (doesn't play well with line numbers), comment out if using method A

%%%%%%%%%%%%%%%  Main text   %%%%%%%%%%%%%%%
% \linenumbers
\section{Introduction}
%With the popularity of online learning and the development of artificial intelligence, smart education, which aims to achieve personalized teaching and learning, has become an important direction in the field of education. Against this backdrop, knowledge tracing(KT), as one of the core technologies of smart education, is committed to capturing students' knowledge mastery state from their historical learning interactions and predicting their learning performance, thereby providing a scientific basis for adaptive learning and precise teaching.

Knowledge Tracing (KT) is a technique that predicts the knowledge state of students by modeling their learning interactions \cite{intro_KT1,intro_KT2}. %In recent years, deep learning-based knowledge tracing has become mainstream and many methods have made remarkable contributions by introducing deep learning techniques\cite{intro_DKT,intro_AKT,intro_Dtransformer}. 
Focusing on the commonalities and characteristics of students in their learning may be more in line with the real scenario and closer to the essence of the KT task \cite{intro_commonalities1,intro_commonalities2}. In view of this, many studies have introduced students' personalized parameters, such as anomalous interaction, emotion, forgetting, etc., in the hope of enhancing the personalized adaptability and interpretability of the KT model. For example, some methods identify ``slip" and ``guess" and eliminate or dilute their effects, thus avoiding misassessing students' knowledge states \cite{intro_DyGKT,intor_ELAKT,intro_UKT,intro_HDKT}. Some methods study the emotional changes of students and the impact of affective states during the problem-solving process \cite{intro_DASKT}. %These studies explain the gap between students' performances and their knowledge mastery states. 
Some methods incorporate student forgetting factors when modeling their learning interactions, achieving this by introducing time intervals \cite{baseline_ReKT}. 
However, these methods treat all students in a common way using only time intervals, ignoring individual differences in forgetting among students. In addition, existing methods ignore the entire memory process that includes forgetting. We all know that memory is an important component of learning ability. Psychologists find it useful to distinguish three stages in the learning/memory process: encoding (acquisition), storage (maintenance or persistence), and retrieval (utilization of stored information) \cite{Melton1963,Weiner1966}. Explicitly modeling this entire process---the encoding of memory at the cognitive level when students learn, the dynamic storage situation of memory before applying knowledge, and the retrieval of memory when utilizing knowledge---is an important task. And forgetting only, as \cite{2018memory} note, exists in the storage part among the three stages. In other words, the existing methods ignore the memory storage stage where forgetting is located and its connection to the encoding and retrieval stages. Therefore, how to jointly model the memory of these three stages is a challenge worth studying for KT.

% However, these methods do the common for all students using only time intervals, ignoring individual differences in forgetting among students. We all know that memory is an important part of learning ability. Psychologists find it useful to distinguish three stages in the learning/memory process: encoding (acquisition), storage (maintenance or persistence), and retrieval (utilization of stored information) \cite{Melton1963,Weiner1966}. While forgetting only exists in the storage part \cite{2018memory}. However, the existing methods ignore the memory storage stage where forgetting is located and its connection to the encoding and retrieval stages. Therefore, how to jointly model the memory of these three stages is a challenge worth studying for KT.

% ignore other memory links as well as personalized forgetting patterns.

% In other words, existing research has not explicitly modeled students' personalized forgetting behavior. We know that memory is an important part of learning ability, and students with strong memory are more likely to learn better. Memory is reflected not only in forgetting knowledge but also in remembering knowledge. However, current KT works are still relatively blank in capturing memory, as existing methods take a default stance on memory. Therefore, in addition to capturing personalized forgetting, how to model memory from students' historical learning records and disentangle it is also a topic worth studying.

Some studies argue that memory is an active generative system rather than passive storage \cite{HemmerS09, FayyazAZKWCW22}. Its essence lies in probabilistic abstraction of the world by the brain/model, leveraging prior knowledge to enable efficient and robust memory reconstruction. Building on this, since the architecture of generative models, particularly Variational Autoencoders (VAE), inherently aligns with the `encoding-storage-retrieval' process of memory, some studies employ VAE to computationally understand and simulate the brain's memory mechanisms \cite{intro_VAEModulateMemory,generativeModelMemory,intro_VAE}. For example, \cite{generativeModelMemory} proposes that trained generative networks capture statistical structures of stored events through integrated memory formation and reproduction. As learning consolidates, such networks support recalling ``facts" and retrieving experiences from these ``facts". Inspired by this, we propose memoryKT to address the aforementioned challenges related to memory. 

Our proposed KT model simulates memory dynamics through a three-stage process. First, it learns the distribution of students' knowledge memory features using a variational encoder. Second, it reconstructs students' history interaction using a variational decoder. Third, it embeds a personalized forgetting algorithm within the LSTM to dynamically modulate memory storage strength. Using a large number of information-rich images as training data, \cite{generativeModelMemory} simulates hippocampal replay with an auto-associative network and trains a generative model using this auto-associative network to mimic collaboration between the hippocampus and the cortex, thus simulating the processes of memory construction and consolidation in the brain. Given that the simple data currently available for KT cannot meet the needs of complex brain-like models, we directly use a VAE to encode the students' knowledge memory features and reconstruct their exercise feedback. To adapt to sequential learning scenarios so as to dynamically simulate memory storage strength, we utilize an LSTM network embedded with a VAE to capture dependencies in the sequence \cite{intro_VRNN}. To distinguish individual differences in students' forgetting behaviors, thereby better modeling the memory storage process, we designed the personalized forgetting algorithm. In this algorithm, we designed a set of points accumulation rules that focus on the time interval and the difficulty of the concept. These rules conform to the laws of forgetting and dynamically update each student's forgetting score based on their interaction records. In theory, a student's grade is positively correlated with his or her memory ability \cite{intro_memoryGradeRelative1,intro_memoryGradeRelative2}, so we designed these rules to obtain a score that better reflects the student's memory capacity. Finally, we used the students' scores to determine their level within the population as a forgetting characteristic, thereby capturing individual differences in the students' forgetting behaviors. In summary, our key contributions are listed below.

%To adapt to sequential learning scenarios, we chose VRNN, which embeds VAE at each time step, to capture dependencies in the sequence. In the personalized forgetting module, we designed the points accumulation rules that conform to the laws of forgetting, dynamically updating the students' forgetting score based on their interactions. For example, when a student encounters the same question or knowledge point again and answers correctly, the longer the interval since the last encounter, the higher the reward. The difficulty of the concept also follows rules similar to the time interval. In theory, a student's grade is positively correlated with their memory ability, so we designed rules that conform to the laws of forgetting to obtain scores that better reflect students' memory capacity. Finally, we use the student's score to determine their level within the population as a forgetting characteristic, thereby capturing individual differences in the student's forgetting. In summary, our key contributions are listed below.

% Our method consists of a VRNN-based memory modeling module to capture and disentangle memory processes in learning sequences\cite{intro_VRNN}, and a personalized forgetting module that uses statistical methods to mine students' personalized forgetting characteristics from data, thus capturing individualized forgetting.
%in the memory modeling module, we directly use a VAE generator to reconstruct interactions for memory modeling, with the encoded random latent variables simulating the student's memory distribution of the interaction, 

\begin{itemize}
\item %For the first time, we introduce memory modeling in knowledge tracing and use the generative model VAE to capture memory.
For the first time, we introduce simulations of the three links of encoding-storage-retrieval in knowledge tracing to model students' memory states, and achieve this goal with an improved temporal variational autoencoder.
\item We explicitly capture students' personalized forgetting in a simple yet effective way and organically integrate it with memory modeling.
\item We conduct comprehensive experiments on four benchmark real-world datasets, and the results demonstrate that memoryKT achieves the best or near-best performance. 

\end{itemize}

\begin{figure*}[htbp] % 使用 figure* 环境使图片跨栏
    \centering
    \includegraphics[width=15cm,height=8cm]{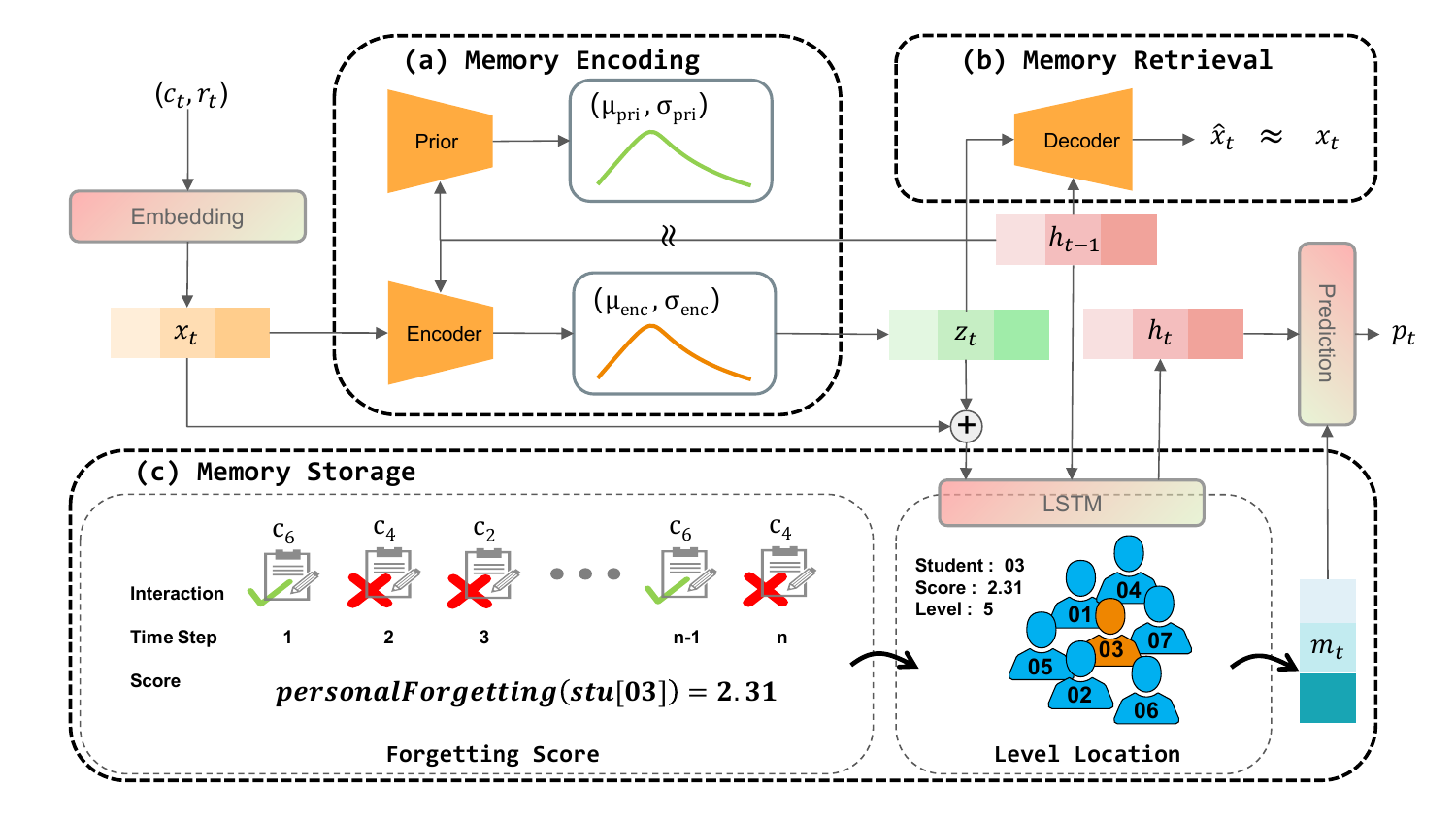} % 图片路径和宽度设置为页面宽度
    \caption{We develop an integrative memory-and-forgetting method for KT. (a) We first utilize the VAE encoder to model the memory encoding while using a prior distribution produced by the historical hidden states to restrain it. (b) Next, the VAE decoder models the memory retrieval link. (c) Then, with the historical hidden states and encoded random states as inputs, we utilize the LSTM to model the memory storage link by capturing the dependencies in the sequence. Finally, the personalized forgetting algorithm is embedded into the temporal workflow. It first dynamically calculates students’ scores based on the laws of forgetting, and then locates their forgetfulness level within the student population.} % 图片说明文字
    \label{The Architecture of memoryKT} % 图片标签，用于引用
\end{figure*}
\section{Problem Definition}
%In this section, we formally define the task of knowledge tracing with memoryKT. In an online learning platform, let $S=\left\{ s_1,s_2,...,s_N \right\}$ denote a set of $N$ students , $C = \left\{c_1,c_2,...,c_K\right\}$ represent a set of $K$ knowledge concepts and $ \varepsilon = \left\{e_1,e_2,...,e_M\right\}$ is a set of $M$ exercises related to these concepts. Suppose that the learning interaction sequence of a student is recorded as $ \mathcal{L}=\left\{(c1,r1,f1),(c2,r2,f2),…,(ct,rt,ft)\right\}$  ,where at the time step $t$: $c_t \in C$ indicates the attempted concept, $r_t \in \left\{0,1\right\}$ records response correctness, and $f_t$ represents real-time forgetting score calculated by the personalized forgetting algorithm. When problem identifiers are available, the interactions can be extended to $(e_t,c_t,r_t,f_t)$. The knowledge tracing task requires dynamically estimating knowledge mastery states and then predicting future performance at the next time step through $p(r_{t+1} = 1 | \mathcal{L},c_{t+1})$, while considering the personalized forgetting effect of the student.
\textbf{Knowledge Tracing}
Let $S = \{s_1, s_2, \dots, s_N\}$ denote the set of students, $C = \{c_1, c_2, \dots, c_K\}$ the set of knowledge concepts, and $\mathcal{E} = \{e_1, e_2, \dots, e_M\}$ the set of exercises. A student's learning interaction sequence is recorded as:
\[
L = \{(c_1, r_1, f_1), (c_2, r_2, f_2), \dots, (c_t, r_t, f_t)\},
\]
where $c_t \in C$ represents the attempted concept, $r_t \in \{0,1\}$ indicates response correctness, and $f_t$ is the real-time forgetting score computed by a personalized forgetting algorithm. The KT task aims to predict:
\begin{equation}
\begin{split}
p(r_{t+1} = 1 \mid L, c_{t+1}).
\end{split}
\end{equation}\\
Three-Stage Memory Modeling is as follows:

% --- Encoding Stage ---
\textbf{Memory Encoding}
When a student interacts with exercise $(e_t, c_t)$ at timestep $t$, the experience is encoded into memory:
\begin{equation}
\begin{split}
\mathbf{h}_t^{\mathrm{enc}} = Enc(h_{t-1},c_t,r_t),
\end{split}
\end{equation}\\
where $\mathrm{Enc}(\cdot)$ is the encoding function, $c_t$ is the concept, $r_t$ is the correctness.

% --- storage Stage ---
\textbf{Storage Stage}
Between timesteps $t$ and $t+1$, the encoded memory decays via a personalized forgetting curve:
\begin{equation}
\begin{split}
\mathbf{h}_t^{\mathrm{sto}} = \mathrm{Sto}(\mathbf{h}_t^{\mathrm{enc}},\Delta\tau,\Phi_s(c_t)),
\end{split}
\end{equation}\\
where $\Delta\tau = \tau_{t+1} - \tau_t$ is the elapsed time, $\Phi_s(c_t)$ captures student $s$'s concept-specific personalized forgetting pattern.

% --- Retrieval Stage ---
\textbf{Memory Retrieval}
At timestep $t+1$, when encountering concept $c_{t+1}$, the student retrieves memory to respond:
\begin{equation}
\begin{split}
p(r_{t+1}=1) = \mathrm{Retrieve}(\mathbf{h}_t^{\mathrm{ret}},c_{t+1},f_{t+1},\mathcal{R}(c_t, c_{t+1})),
\end{split}
\end{equation}\\
where $\mathcal{R}(c_t, c_{t+1})$ denotes the conceptual relatedness, $f_{t+1}$ is the real-time forgetting.

\section{Methodology}

Inspired by the research of \cite{generativeModelMemory}, we use the VAE to capture the memory processes of students during online learning. To dynamically simulate the memory storage strength, we propose the memoryKT approach based on LSTM. We also design the personalized forgetting module, which calculates the personalized forgetting property to explicitly model the students' personalized forgetting curve. Fig.~\ref{The Architecture of memoryKT} presents the framework of the memoryKT approach. In this section, we will elaborate on the specific modules of memoryKT.

\subsection{Embedding}
Our method focuses on memory modeling, so for the input of the memory process, memoryKT can adapt the embeddings from many other methods. Although embeddings closer to real-world interactions are theoretically preferable, to process datasets containing only knowledge concepts without explicit question representations (e.g., AS2015), the embedding layer employs a minimal design utilizing knowledge concepts and student responses. Formally, the embedding at the time step $t$ is computed as follows:

\begin{equation}\label{eq5}
    x_t=Embedding(c_t+K\cdot r_t),
\end{equation}
 where $c_t$ denotes the knowledge concept identifier, $r_t$ indicates the correctness of the response, and $K$ represents the number of the knowledge concept.

\subsection{Memory Encoding}
\subsubsection{Encoder}
Similar to how the brain stores memories, the encoder employs a neural network to encode the embedding of the student's interaction $x_t$ and the previous hidden state $h_{t-1} $ into the mean $\mu_{enc}$ and the standard deviation $\sigma_{enc}$. We consider it as the student's memory distribution concerning the relevant interaction. The procedure involves first subjecting the input embedding to nonlinear feature extraction via two fully connected layers equipped with ReLU activation functions. Following this, the resulting feature vector is concatenated with the previous hidden state $h_{t-1} $ and further processed through an encoder network consisting of two linear layers with ReLU. Eventually, two distinct linear layers produce the mean $\mu_{enc}$  and standard deviation $\sigma_{enc}$  of the variational posterior distribution, with the standard deviation ensured non-negative by the $Softplus$ activation function. 

\begin{equation}
\begin{split}
\phi_x^t &= ReLU(W_2\cdot ReLU(W_1\cdot x_t+b_1)+b_2),\\
enc_t&=ReLU(W_e\cdot [\phi_x^t;h_{t-1}]+b_e),\\
\mu_{enc}^t&=W_{\mu}\cdot enc_t,\quad \sigma_{enc}^t=Softplus(W_{\sigma}\cdot enc_t),
\end{split}
\end{equation}
where $h_{t-1}$ is the hidden state of the $t-1$ timestep. $W_1$, $W_2$, $b_1$, and $b_2$ denote the trainable weights and biases of the input embedding processing layers. $W_e$ and $b_e$ are the trainable weights and biases of the encoder network, and $W_\mu$ and $W_\sigma$ are the trainable weight matrices of the linear layers that generate the mean and standard deviation.

\subsubsection{Prior Distribution}
Unlike VAE's assumption that the prior distribution is a standard normal distribution, our prior distribution derived from $h_{t-1}$ is employed to constrain the posterior distribution obtained during the encoding phase to introduce temporal dependencies and stochasticity. Specifically, the prior distribution is modeled as follows:

\begin{equation}
\begin{split}
    prior_t&=ReLU(W_p\cdot h_{t-1}+b_p),\\ \mu_{prior}^t&=W_{\mu_p}\cdot prior_t,  \sigma_{prior}^t=Softplus(W_{\sigma_p}\cdot prior_t),
\end{split}
\end{equation}
% where $h_{t-1}$ is the hidden state of the t-1 timestep, which
where the previous hidden state $h_{t-1}$ is processed by a prior network with a $ReLU$ activation function; two different linear layers produce the mean of the prior distribution $\mu_{prior}$ and the standard deviation $\sigma_{prior}$ respectively, $Softplus$ is similarly used to ensure the non-negativity of the standard deviation $\sigma_{prior}$.%It takes the previous hidden state $h_{t-1}$ is the previous hidden state, it is processed through a prior network with a ReLU activation function, two distinct linear layers separately produce the mean $\mu_{prior}$ and standard deviation $\sigma_{prior}$ of the prior distribution, and $Softplus$ is similarly used to ensure the non-negativity of the standard deviation $\sigma_{prior}$. 
$W_p$ and $b_p$ denote the trainable weights and biases of the prior network, while $W_{\mu_p}$ and $W_{\sigma_p}$ represent the trainable weight matrices to generate the mean and standard deviation of the prior distribution.

\subsection{Memory Retrieval}
The decoder simulates the student's recall process by utilizing the memory stochastic distribution and hidden state as input, modeling how students retrieve relevant memories and temporal information for recall. To enable end-to-end training while maintaining differentiability, the reparameterization trick is employed to sample the learned memory distribution. Specifically, the reparameterization technique transforms the sampling operation into a deterministic function; the latent memory variable $z_{t}$ is obtained through the first formula of Eq. \ref{eq8}, where $\epsilon$ is sampled from a standard normal distribution. The sampled latent variable $z_{t}$ is subsequently processed through a single-layer ReLU-activated feature extraction layer. The resulting features $\phi_z^{t}$ are then concatenated with the previous hidden state $h_{t-1}$ and processed by two fully connected layers with the ReLU function. Finally, the final reconstruction $\widehat{x}_t$ is obtained through a linear layer with a sigmoid activation function. 
\begin{equation}\label{eq8}
\begin{split}
    z_t &= \mu_{enc}+\sigma_{enc}\odot \epsilon, \epsilon \sim \mathcal{N}(0,1),\\
    \phi_z^t &= ReLU(W_z\cdot z_t+b_z),\\
    dec_t&=ReLU(W_d\cdot [\phi_z^t;h_{t-1}]+b_d),\\
    \widehat{x}_t&=\sigma(W_{out}\cdot dec_t),
\end{split}
\end{equation}
where $W_z$, $b_z$, $W_d$, $b_d$, and $W_{out}$ are trainable parameters, and $\sigma(\cdot)$ is the sigmoid activation function.
\subsection{Memory Storage}
% The memory encoding step and retrieval step capture the process of students memorizing interaction, while the personalized forgetting module accounts for students' forgetting of memories.
\subsubsection{Update}
To modulate the memory storage dynamics, the model uses an LSTM to update the hidden state over time. At each timestep, the LSTM receives the concatenation of the input features $\phi_x^t$ and the latent features $\phi_z^t$, along with the previous hidden state $h_{t-1}$ and the cell state ${cell}_{t-1}$.   
\begin{equation}
    \begin{split}
        (h_t,cell_t) &= LSTM([\phi_x^t;\phi_z^t],(h_{t-1},cell_{t-1})),
    \end{split}
\end{equation}
where LSTM stands for long short-term memory. This gated mechanism enables the model to capture long-range dependencies while mitigating vanishing gradient issues.
\subsubsection{Personal Forgetting}

 Different students exhibit different forgetting rates, and thus applying an identical time-interval-based forgetting treatment to all students neglects individual characteristics. We explicitly capture personalized forgetting by statistically computing each student's forgetting characteristics from their historical behavioral data. 

As shown in Algorithm~\ref{algorithmPersonalForget}, the preprocessing phase first extracts knowledge concept difficulty and cohort-level performance statistics. The former serves as a difficulty weight in calculating the student's forgetting characteristic score, while the latter positions the student within the cohort to evaluate the student's forgetfulness level. During the processing of the current interaction, we update the student's forgetting characteristic score using a method that conforms to forgetting patterns, based on the correctness of the current answer, the time interval since the last related question, and the difficulty of the question. Specifically, when the current question is answered correctly, a longer time interval since the last related question yields a larger score increase, and higher question difficulty similarly yields a larger score increase. Conversely, when the current question is answered incorrectly, a shorter time interval since the last related question results in a larger score decrease, and lower question difficulty results in a larger score decrease, particularly when the last related question was answered correctly. To mitigate the impact of the number of answers on the forgetting characteristic score, the score adjustment is normalized by the total number of responses. After updating the current forgetting characteristic score, we project it onto the cohort score distribution and assess the student's forgetfulness level based on his position within this cohort.

\begin{algorithm}
\caption{Personalized Forgetting Level Calculation}
\label{algorithmPersonalForget}
\begin{algorithmic}[1]
\State \textbf{Input:} Student sequence $ S = \{(c_i, r_i, t_i)\}_{i=1}^T $
\Statex Pre-computed concept difficulties $ \mathcal{D} = \{d_1, \dots, d_N\} $
\Statex Pre-computed population score distribution $ \mathcal{P}_{scores} $
\State \textbf{Output:} Forgetting level $ L \in [1, 10] $
\Function{$getForgetLevel$}{$S, D, \mathcal{P}_{scores}$}
    \State $ \text{stu\_score} \gets \text{mean}(\mathcal{P}_{scores}) $ 
    \State \Comment{Cold-start with population mean}
    \State $ \text{last\_interaction} \gets \{\} $
    \State\Comment{Track last interaction for each concept}
    \For{$i = 1 $ to $ T $}
        \State $ (c_i, r_i, t_i) \gets S[i] $
        \State $ (t_{prev}, r_{prev})$
        \State $\gets\text{last\_interaction.get}(c_i,(t_i,\text{None}))$
        \State $ \Delta t \gets t_i - t_{prev} $
        \State \Comment{Time interval since last interaction}
        \State $ \Delta S \gets \text{ScoreChange}(r_i, d_{c_i}, \Delta t, r_{prev}) $
        \State $\text{stu\_score} \gets \text{stu\_score} + \Delta S$
        \State \Comment{Update score}
        \State $ \text{last\_interaction}[c_i] \gets (t_i, r_i) $
    \EndFor
    \State $ L \gets \text{GetPercentile}(\text{stu\_score},\mathcal{P}_{scores}) $
    \State \Comment{Map percentile to level [1, 10]}
    \State \textbf{return} $ L $
\EndFunction
\end{algorithmic}
\end{algorithm}
%wait to update
% \begin{algorithm}
% \caption{Personalized Forgetting Level Computation}
% \begin{algorithmic}[1]
% \State \textbf{Input:} Student sequence \( S = \{(c_i, r_i, t_i)\}_{i=1}^T \)
% \State \textbf{Output:} Forgetting level \( L \in [1, 10] \)
% \State \textbf{Initialize:} \( \text{score} \leftarrow 0 \), \( \text{count} \leftarrow 0 \), \( \text{last} \leftarrow \{\} \)

% \For{\( i = 1 \) to \( T \)}
%     \State \( c, r, t \leftarrow S[i] \)
%     \If{\( c \in \text{last} \)}
%         \State \( \Delta t \leftarrow t - \text{last}[c].\text{time} \)
%         \State \( r_{\text{prev}} \leftarrow \text{last}[c].\text{response} \)
%     \Else
%         \State \( \Delta t \leftarrow 1 \), \( r_{\text{prev}} \leftarrow \text{None} \)
%     \EndIf
    
%     \State \( d \leftarrow \text{difficulty}[c] \)
%     \If{\( r = 1 \)} \textbf{// Correct}
%         \State \( \text{score} \leftarrow \text{score} + \Delta t \times d \times \beta \)
%     \Else \textbf{// Incorrect}
%         \State \( \text{penalty} \leftarrow \frac{(1-d) \times \beta}{\Delta t} \)
%         \If{\( r_{\text{prev}} = 1 \)}
%             \State \( \text{penalty} \leftarrow 1.5 \times \text{penalty} \)
%         \EndIf
%         \State \( \text{score} \leftarrow \text{score} - \text{penalty} \)
%     \EndIf
    
%     \State \( \text{last}[c] \leftarrow (t, r) \), \( \text{count} \leftarrow \text{count} + 1 \)
% \EndFor

% \State \( L \leftarrow \text{ProjectToCohort}(\frac{\text{score}}{\text{count}}, H) \)
% \State \textbf{return} \( L \)
% \end{algorithmic}
% \end{algorithm}

\subsection{Prediction Layer}
Upon obtaining the student's personalized forgetting characteristics, these features are integrated with the hidden state and time interval as input to the prediction layer to capture student forgetting patterns. The embedding of the personalized forgetting level $m_t$ is extracted using the personalized forgetting algorithm. Feature fusion and compression are performed by concatenating the hidden state $h_t$, the personalized forgetting level $m_t$, and the time interval $\Delta t$, which are then processed through a two-layer feedforward network with the ReLU activation function. Finally, the prediction of the student's performance over the concepts is obtained by applying a sigmoid activation function.

\begin{equation}
    \begin{split}
        m_t &= personalForget(x_t),\\
        f_t&=ReLU(W_{f2}\cdot ReLU(W_{f1}\cdot \left[
	\begin{array}{ccc}
		h_t \\
		m_t  \\
		\Delta t
	\end{array}
        \right]+b_{f1})+b_{f2}),\\
        p_t&=\sigma(W_{f3}\cdot f_t+b_{f3}),
    \end{split}
\end{equation}
where $W_{f1}$, $W_{f2}$, and $W_{f3}$ represent the trainable weight matrices, and $b_{f1}$, $b_{f2}$, and $b_{f3}$ denote the corresponding bias vectors.

\subsection{Model Optimization }
To train all the parameters, the memoryKT model is optimized through a multiobjective loss function that balances reconstruction effect, latent space regularization, and prediction accuracy. The optimization framework comprises three primary components:

%First, the reconstruction loss \\$ \mathcal{L} _{recon}$ quantifies the discrepancy between the reconstructed input $\hat{x}$ and the original input $x$ across all time steps, formulated as: 
\begin{equation}
    \begin{split}
        \mathcal{L}_{recon} &= \frac{1}{T}\sum\limits_{t=1}^T{\lvert \lvert \hat{x}_t-x_t\rvert\rvert}^2,
    \end{split}
\end{equation}

%Second, the Kullback-Leibler divergence loss $\mathcal{L}_{KL}$ enforces temporal consistency in the latent space by constraining the posterior distribution $q(z_t|x_t,h_{t-1})$ to align with the prior distribution $p(z_t|h_{t-1})$:

\begin{equation}
    \begin{split}
        \mathcal{L}_{KL} &= \frac{1}{T}\sum\limits_{t=1}^T{D_{KL}(q(z_t\vert x_t,h_{t-1})\vert\vert p(z_t\vert h_{t-1}))},
    \end{split}
\end{equation}

%Third, the prediction loss $\mathcal{L}_{pred}$ evaluates the model's ability to forecast student responses using binary cross-entropy between the predicted probability $p_t$ and the actual response $r_{t+1}$, with masking to handle missing data: 

\begin{equation}
    \begin{split}
        \mathcal{L}_{pred} &= \frac{1}{T-1}\sum\limits_{t=1}^{T-1}BCE(p_t,r_{t+1})\odot mask_{t+1}.
    \end{split}
\end{equation}

The total loss $\mathcal{L}$ is the sum of these three components, providing a balanced optimization objective that simultaneously enhances reconstruction quality, maintains temporal coherence in the latent space, and improves prediction accuracy.  The Adam optimizer is used to minimize the objective function. $\lambda_{rec}$ and $\lambda_{kld}$ are the hyperparameters.

\begin{equation}
    \begin{split}
        \mathcal{L}&=\lambda_{rec}\cdot\mathcal{L}_{recon}+\lambda_{kld}\cdot\mathcal{L}_{KL}+\mathcal{L}_{pred}.
    \end{split}
\end{equation}

\section{Experiments}
In this section, we first introduce the experimental setup and then conduct extensive experiments to compare the performance of the proposed memoryKT with other models, evaluate the contribution of the modules within memoryKT, analyze the hyperparameters, and analyze memoryKT's ability to disentangle memory.

% \begin{table*}[ht]
% \centering
% \adjustbox{max width=\textwidth}{
% \begin{tabular}{@{}l c *{7}{c} @{}}
% \toprule
% \multirow{2}{*}{\textbf{Method}} & \multicolumn{3}{c}{AUC} & \multicolumn{3}{c}{ACC}  \\
% \cmidrule(lr){2-4} \cmidrule(lr){5-7} 
% &ASSIST09 & ASSIST15 & ???? & ASSIST09 & ASSIST15 & POJ \\
% \midrule
% % FPMC  &SIGIR'16   & 0.1003 & 0.2126 & 0.2970 & 0.3423 & 0.1701 & 0.0814 & 0.2045 & 0.2746 & 0.3450 & 0.1344  \\
% DKT   &$0.8226\pm0.0000$  &$0.7271\pm0.0000 $ & $0.0000\pm0.0000$ & $0.7657\pm0.0000$ & $0.7503\pm0.0000$ & $0.0000\pm0.0000$ \\
% DKVMN &0.7330 & 0.7227 & 0.0000 & 0.7144 & 0.7508 & 0.0000 \\
% GKT   &0.7227 & 0.7258 & 0.0000 & 0.7077 & 0.7504 & 0.0000 \\
% SAKT  &0.7746 & 0.7114 & 0.0000 & 0.7063 & 0.7474 & 0.0000 \\
% AKT   &0.7650 & 0.7281 & 0.0000 & 0.7323 & 0.7521 & 0.0000 \\
% SAINT &0.7458 & 0.7026 & 0.0000 & 0.6936 & 0.7438 & 0.0000 \\
% sparseKT  & 0 & 0.7219 & 0.0000 & 0 &0.7499 & 0.0000 \\
% simpleKT  &0.8413 &0.7248 & 0.0000 &0.7748 &0.7508 & 0.0000 \\
% UKT  &0.8563 &0.7267 & 0.0000 &\underline{0.7814} &\underline{0.7497} & 0.0000 \\
% \midrule
% \textbf{Ours} & \textbf{0.8534} & \textbf{0.7455} & \textbf{0.0000} & \textbf{0.7965} & \textbf{0.7582} & \textbf{0.0000} \\
% \bottomrule
% \end{tabular}
% }
% \caption{Overall AUC and Accuracy performance of memoryKT and all baselines.}
% \label{tab:Overall Performance}
% \end{table*}

\subsection{Experiential Settings}
\subsubsection{Datasets}
To evaluate the effectiveness of memoryKT, we select four widely used real-world datasets ASSIST09, ASSIST15, AL2005, and POJ.
% \footnote{\url{https://sites.google.com/site/assistmentsdata/home/20092010-assistment-data.}}
% \footnote{\url{https://sites.google.com/site/assistmentsdata/datasets/2015assistments-skill-builder-data.}}
% \footnote{\url{https://pslcdatashop.web.cmu.edu/KDDCup/.}}
% \footnote{\url{https://drive.google.com/drive/folders/1LRljqWfODwTYRMPw6wEJ_mMt1KZ4xBDk.}}
\begin{itemize}
    \item \textbf{ASSIST09} This dataset comprises mathematical exercises collected from the free online tutoring platform ASSISTments during the 2009-2010 academic year \cite{dataset_ASSIST09}. It has been widely used for KT methods in the past decade. The dataset contains 337,415 learning interactions across 4,661 student sequences, encompassing 17,737 unique questions and 123 knowledge components (KCs). Each question is associated with an average of 1.1968 KCs.
    \item \textbf{ASSIST15} Similar to ASSIST09, this dataset is collected from the ASSISTments platform during the 2015 academic year. It distinguishes itself by having the largest student population among all ASSISTments datasets, making it particularly valuable for large-scale educational analytics research. The dataset comprises 682,789 learning interactions distributed across 19,292 student sequences, with 100 distinct knowledge components.
    \item \textbf{AL2005} Originating from the KDD Cup 2010 EDM Challenge \cite{dataset_AL2005}, this dataset contains step-level responses of 13-14 year-old students to Algebra problems. It features 884,098 learning interactions across 4,712 sequences, with 173,113 unique questions constructed through problem-step concatenation and 112 knowledge components (KCs), averaging 1.3521 KCs per question.
    \item \textbf{POJ} This programming exercise dataset is collected from Peking University's online coding practice platform, originally scraped by \cite{dataset_POJ}. It contains 884,089 learning interactions across 20,114 student sequences, featuring 2,748 unique programming problems.
\end{itemize}
The statistics of the four datasets are listed in Tab.~\ref{tab:dataset_stats}.
% The statistics of the three datasets are listed in Table 2.

\begin{table}[H]
\footnotesize          % 与 A 表一致
\centering
\begin{tabular}{@{}l c c c c@{}}   % 与 A 表对齐方式一致
\toprule
\textbf{Dataset} & \textbf{ASSIST09} & \textbf{ASSIST15} & \textbf{AL2005} & \textbf{POJ} \\
\midrule
\#Student      & 4,160  & 19,292 & 574   & 20,114 \\
\#Interaction  & 337,415& 682,789& 987,593& 884,089\\
\#Sequences    & 4,661  & 19,292 & 4,712 & 20,114\\
\#Question     & 17,737 & \textbf{--} & 173,113 & 2,748\\
\#Concept      & 123    & 100   & 112   & \textbf{--}\\
\bottomrule
\end{tabular}
\caption{Dataset statistics}
\label{tab:dataset_stats}
\end{table}

% \begin{table}[H]
% \centering
% \setlength{\tabcolsep}{3pt}
% \begin{tabular}{l c c c c c}
% \toprule
%  \textbf{Dataset} & \textbf{ASSIST09} & \textbf{ASSIST15} & \textbf{AL2005}& \textbf{POJ} \\
% \midrule
% \#Student &4,160 &19,292  &574 &20,114  \\
% \#Interaction &337,415 &682,789 &987,593 &884,089 \\
% \#Sequences&4,661&19,292&4,712&20,114\\
% \#Question &17,737&\textbf{-}&173,113 &2,748\\
% \#Concept &123&100&112 &\textbf{-}\\
% \bottomrule
% \end{tabular}
% \caption{Dataset statistics}
% \label{tab:dataset_stats}
% \end{table}

\subsubsection{Baselines}
We evaluated the performance of the proposed memoryKT by comparing it with ten powerful and widely selected baselines. Including DKT \cite{intro_DKT}, DKVMN \cite{baseline_DKVMN}, SAKT \cite{baseline_SAKT}, GKT \cite{baseline_GKT}, KQN \cite{baseline_KQN}, SAINT \cite{baseline_SAINT}, AKT \cite{intro_AKT}, ATKT \cite{baseline_ATKT}, simpleKT \cite{baseline_simpleKT}, and ReKT \cite{baseline_ReKT}. The specific details of the aforementioned baselines are listed in the appendix.

\subsubsection{Implementation Details}
To fairly compare with other methods, we embedded memoryKT into the PyKT \cite{exp_PyKT} environment to employ a unified approach for training and evaluating all models. Specifically, students with fewer than 3 interactions were filtered out, 20\% of the data was randomly selected as the test set, and the remaining 80\% was randomly divided into five parts for 5-fold cross-validation. The models were trained up to 200 epochs using the Adam optimizer with an early stopping strategy. The embedding dimension is set to 64, the hidden state dimension to 128, and the latent variable dimension to 32. Learning rates are set to 1e-3, dropout rates to 0.1, with weight decay of 1e-5. Our model is implemented with PyTorch and trained on a single A100 GPU. To be in line with previous works, we use AUC (Area Under the Curve) as the first evaluation metric and ACC (Accuracy) as the second metric.

\subsection{Results}

\begin{table*}[ht] 
\footnotesize
 \centering 
 \adjustbox{max width=\textwidth}{ 
 \begin{tabular}{@{}l c *{8}{c} @{}} 
 \toprule 
 \multirow{2}{*}{\textbf{Method}} & \multicolumn{4}{c}{AUC} & \multicolumn{4}{c}{ACC}  \\ 
 \cmidrule(lr){2-5} \cmidrule(lr){6-9} 
 &ASSIST09 & ASSIST15 & AL2005 & POJ & ASSIST09 & ASSIST15 & AL2005 & POJ \\ 
 \midrule 
 DKT   &0.8226 &0.7271 &0.8146 &0.6089 &0.7657 &0.7503 &0.7882 &0.6328 \\ 
 DKVMN &0.8213 &0.7227 &0.7891 &0.6056 &0.7650 &0.7508 &0.7778 &0.6393 \\  
 SAKT  &0.7746 &0.7114 &0.7682 &0.6095 &0.7063 &0.7474 &0.7729 &0.6407 \\ 
 GKT   &0.8171 &0.7258 &0.8025 &0.6070 &0.7609 &0.7504 &0.7825 &0.6117 \\
 KQN   &0.8216 &0.7254 &0.8005 &0.6080 &0.7659 &0.7500 &0.7850 &0.6435 \\ 
 SAINT &0.7458 &0.7026 &0.6662 &0.5563 &0.6936 &0.7438 &0.7538 &0.6467 \\ 
 AKT   &\underline{0.8474} &\underline{0.7281} &0.8091 &\textbf{0.6281} &\underline{0.7772} &\underline{0.7521} &0.7939 &0.6492 \\ 
 ATKT  &0.7337 &0.7245 &0.7964 &0.6075 &0.7208 &0.7494 &0.7774 &0.6332 \\ 
 simpleKT &0.8413 &0.7248 &0.8254 &0.6194 &0.7748 &0.7508 &0.8083 &\underline{0.6498} \\ 
 ReKT &0.7861 &0.7243 &\underline{0.8275} &\underline{0.6272} &0.7364 &0.7513 &\textbf{0.8164} &\textbf{0.6553}  \\ 
 \midrule 
 \textbf{memoryKT} &\textbf{0.8534} &\textbf{0.7476} &\textbf{0.8437} &0.6186 &\textbf{0.7974} &\textbf{0.7590} &\underline{0.8117} &0.6473 \\ 
 % \textbf{memoryKT} &\textbf{0.8534} &\textbf{0.7447} &\textbf{0.8355} &0.6125 &\textbf{0.7974} &\textbf{0.7574} &0.8066 &0.6406 \\ 
 \bottomrule 
 \end{tabular} 
 } 
 \caption{Overall AUC and Accuracy performance of memoryKT and all baselines.} 
 \label{tab:Overall Performance} 
 \end{table*}

\subsubsection{Overall Performance}
Evaluated with AUC and ACC metrics, Tab.~\ref{tab:Overall Performance} presents the performance comparison in four datasets between memoryKT and other advanced methods covering major deep learning techniques, with the best results highlighted in bold and the second-best results underlined. According to Tab.~\ref{tab:Overall Performance}, we can observe the following:
 (1) Overall, memoryKT significantly outperforms other baselines. In datasets other than POJ, memoryKT only falls behind the advanced ReKT by a marginal difference of 0.0047 in ACC on the AL2005 dataset. In POJ, memoryKT closely follows simpleKT, ReKT, and AKT, ranking fourth in the first tier and leading other baselines. It can be said that memoryKT achieves the best or near-best performance.
 (2) We found that the POJ dataset contains only problem information without concept information, and the embedding layer employed by memoryKT is not suitable for this scenario. Meanwhile, the better-performing simpleKT and AKT adopt the attention mechanism, and ReKT, which models directly on the fine-grained problem sequence, also achieves an effect similar to the attention mechanism. This may suggest that the attention mechanism is better suited to the POJ dataset.
 (3) Unlike traditional approaches that primarily focus on the interaction between students and knowledge at the behavioral level (response correctness and response sequences), memoryKT models the interaction between students and knowledge at the cognitive level (memory encoding, storage, and retrieval). Although memoryKT does not always achieve the best results on all metrics across all datasets, compared with other methods, it is able to capture finer-grained changes in knowledge mastery and to model individualized forgetting patterns more effectively.

\subsubsection{Ablation Study}
\begin{figure}[htbp] 
   \centering 
   % ---------- 第一行：图例 ---------- 
   \begin{subfigure}{\linewidth}  % 宽度占满整行 
     \centering 
     \includegraphics[width=\linewidth]{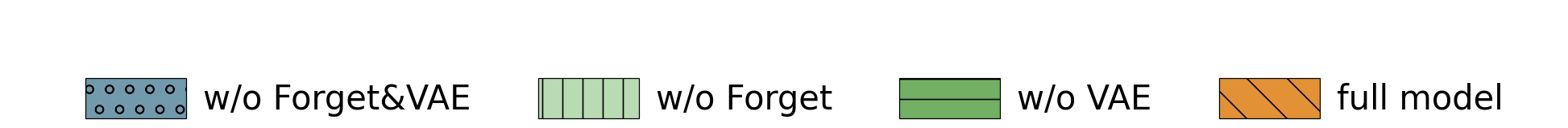}  
     \label{fig:legend} 
   \end{subfigure} 
   % \par\smallskip  % 竖直间距，可按需调整 
   
   % ---------- 第二行：ASSIST09数据集 ---------- 
   \begin{subfigure}{0.48\linewidth} 
     \centering 
     \includegraphics[width=\linewidth]{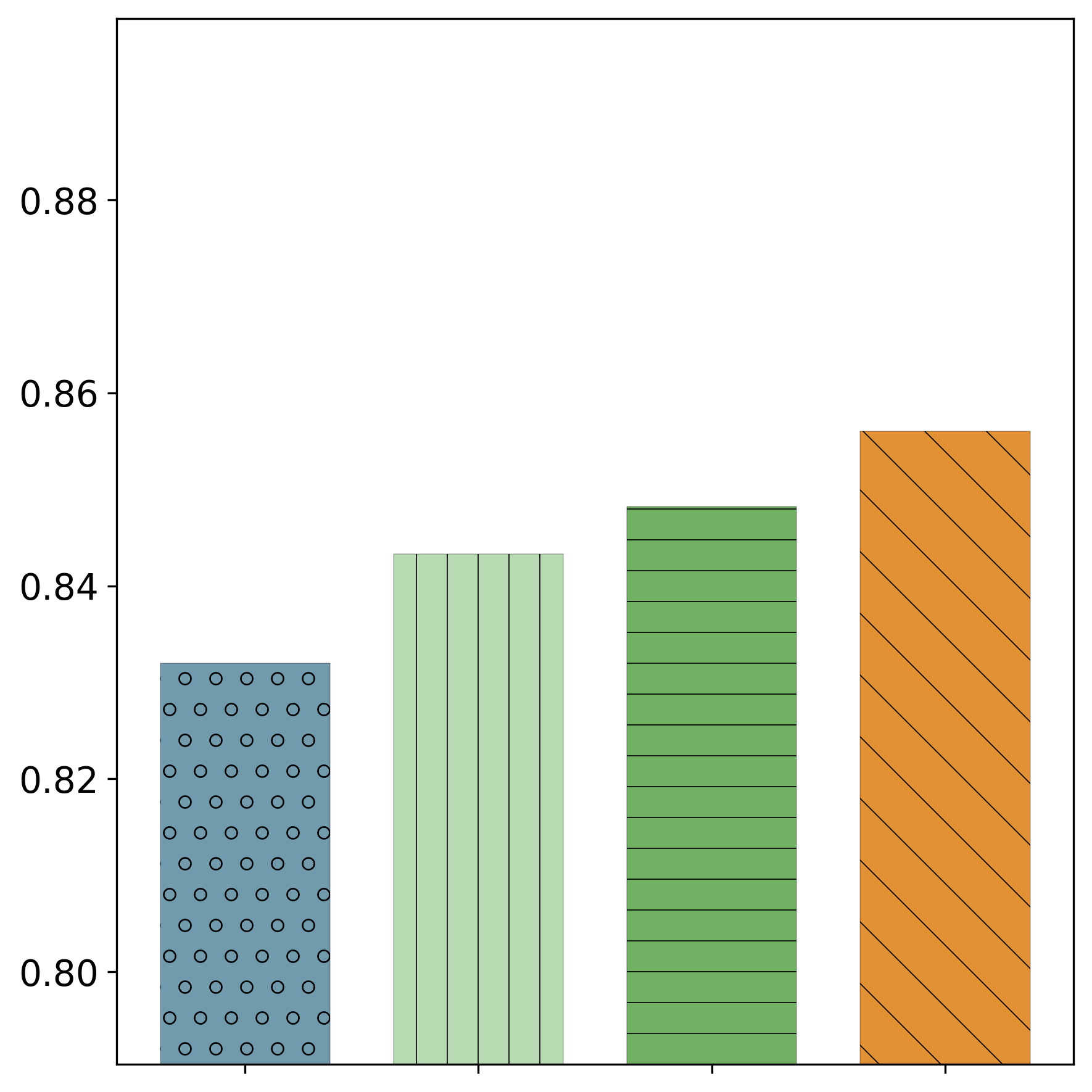} 
     \subcaption{AUC on ASSIST09 dataset} 
     \label{fig:as09-auc} 
   \end{subfigure} 
   \hfill 
   \begin{subfigure}{0.48\linewidth} 
     \centering 
     \includegraphics[width=\linewidth]{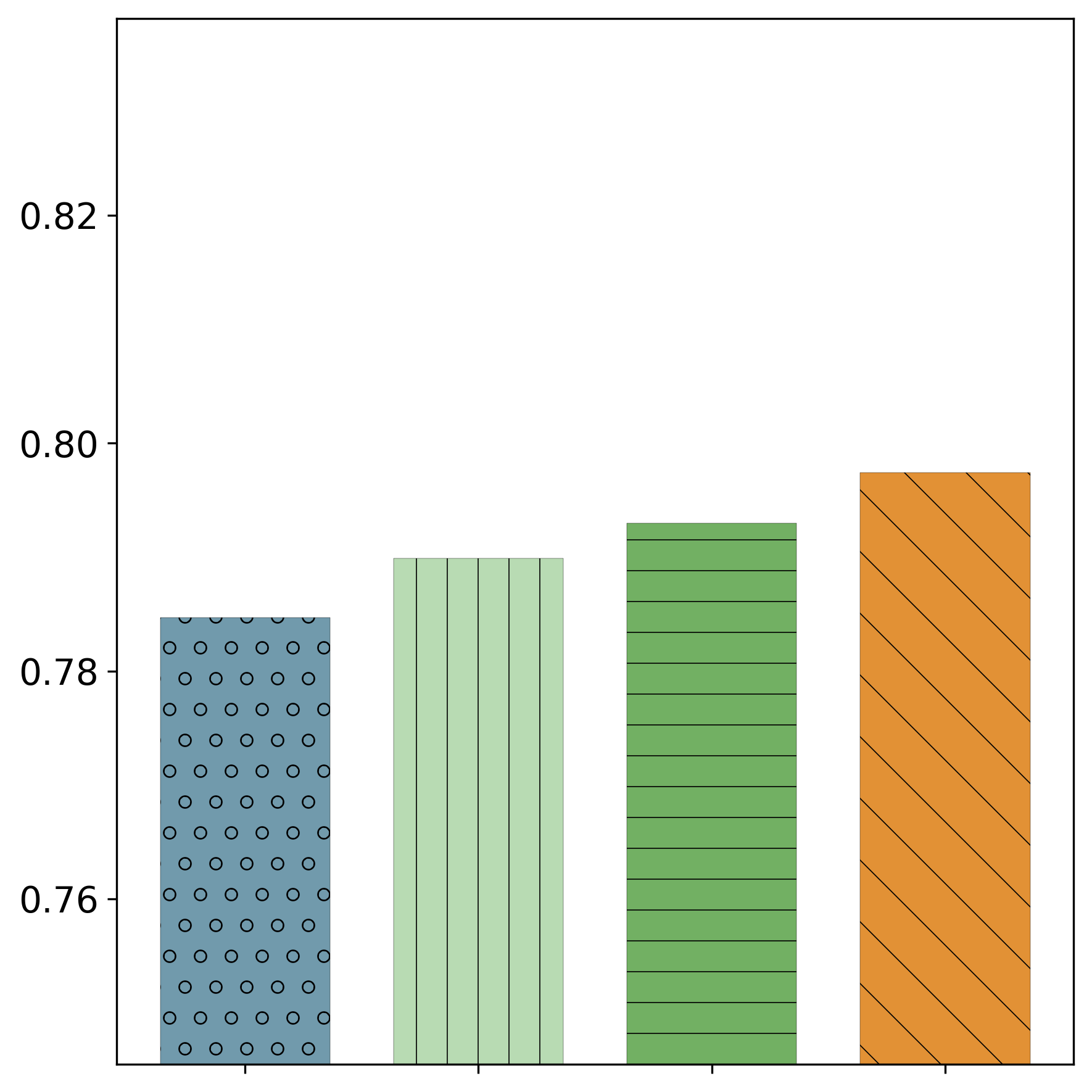} 
     \subcaption{ACC on ASSIST09 dataset} 
     \label{fig:as09-acc} 
   \end{subfigure} 
   \par\bigskip  % 竖直间距，可按需调整 
   
   % ---------- 第三行：ASSIST15数据集 ---------- 
   \begin{subfigure}{0.48\linewidth} 
     \centering 
     \includegraphics[width=\linewidth]{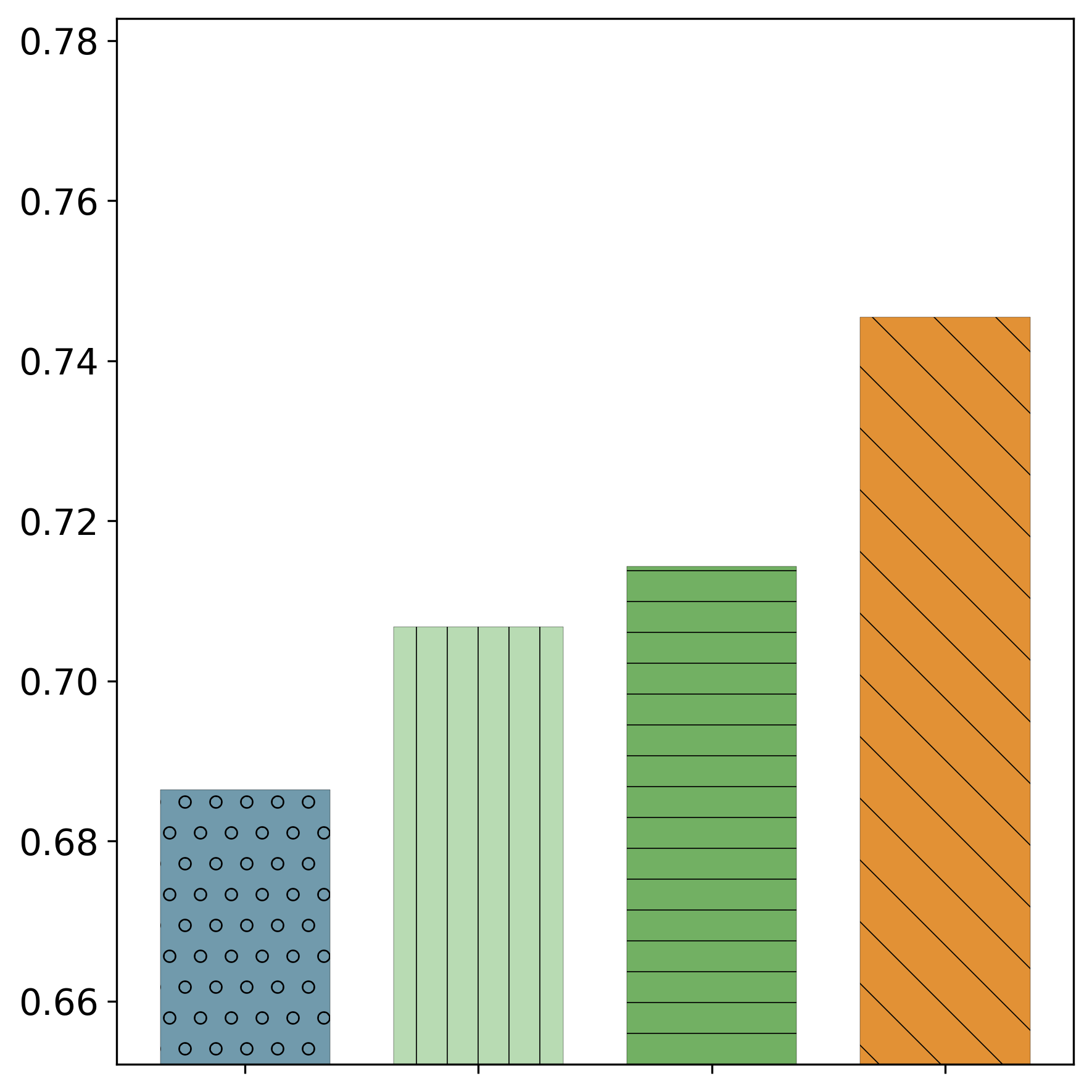} 
     \subcaption{AUC on ASSIST15 dataset} 
     \label{fig:as15-auc} 
   \end{subfigure} 
   \hfill 
   \begin{subfigure}{0.48\linewidth} 
     \centering 
     \includegraphics[width=\linewidth]{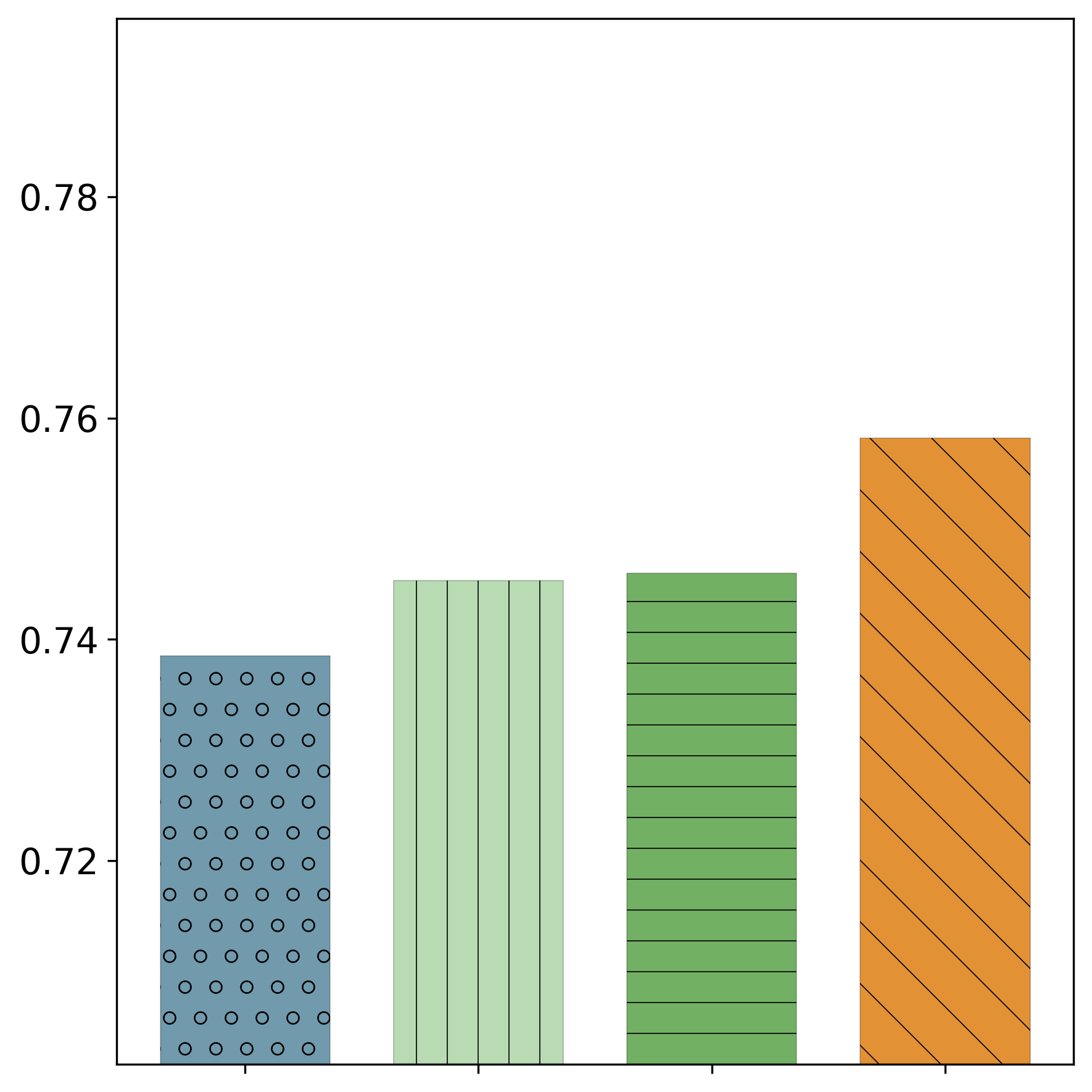} 
     \subcaption{ACC on ASSIST15 dataset} 
     \label{fig:as15-acc} 
   \end{subfigure} 
 
   \caption{Ablation study} 
   \label{Ablation Study} 
 \end{figure}

To explore the contribution of different components to memoryKT, we constructed three variants of the model to compare with the original model on datasets ASSIST09 and ASSIST15, as shown in Fig.~ \ref{Ablation Study}. Specifically, ``w/o VAE" removes the memory encoding and decoding core, the VAE; ``w/o forget" removes the personal forgetting module; and ``w/o forget\&VAE" removes both of the above from the proposed memoryKT. From Fig.~\ref{Ablation Study}, we have the following observations: (1) Both ``w/o VAE" and ``w/o forget" exhibit significant performance degradation compared to the original model, particularly in the AUC metric, indicating that both components are indispensable for the overall framework. (2) The ``w/o VAE" and ``w/o forget" demonstrate comparable performance, with ``w/o VAE" marginally outperforming the latter. We speculate that among the three links of memory, the memory storage link accounts for a heavier weight, and the personalized forgetting algorithm is indispensable to model this process. (3) As anticipated, the ``w/o forget\&VAE" variant yields the most substantial performance decline. This observation implies that the model synergistically integrates the mechanisms of memory encoding, retrieval, and storage. (4) The consistency of the distribution of variants across two datasets and two metrics corroborates the contribution of each component to the whole.

\begin{figure}[htbp]
  \centering
  \scalebox{0.999}{
        % ---------- 子图 a ----------
      \begin{subfigure}{1\linewidth}  % 宽度占满整行
        \centering
        \includegraphics[width=\linewidth]{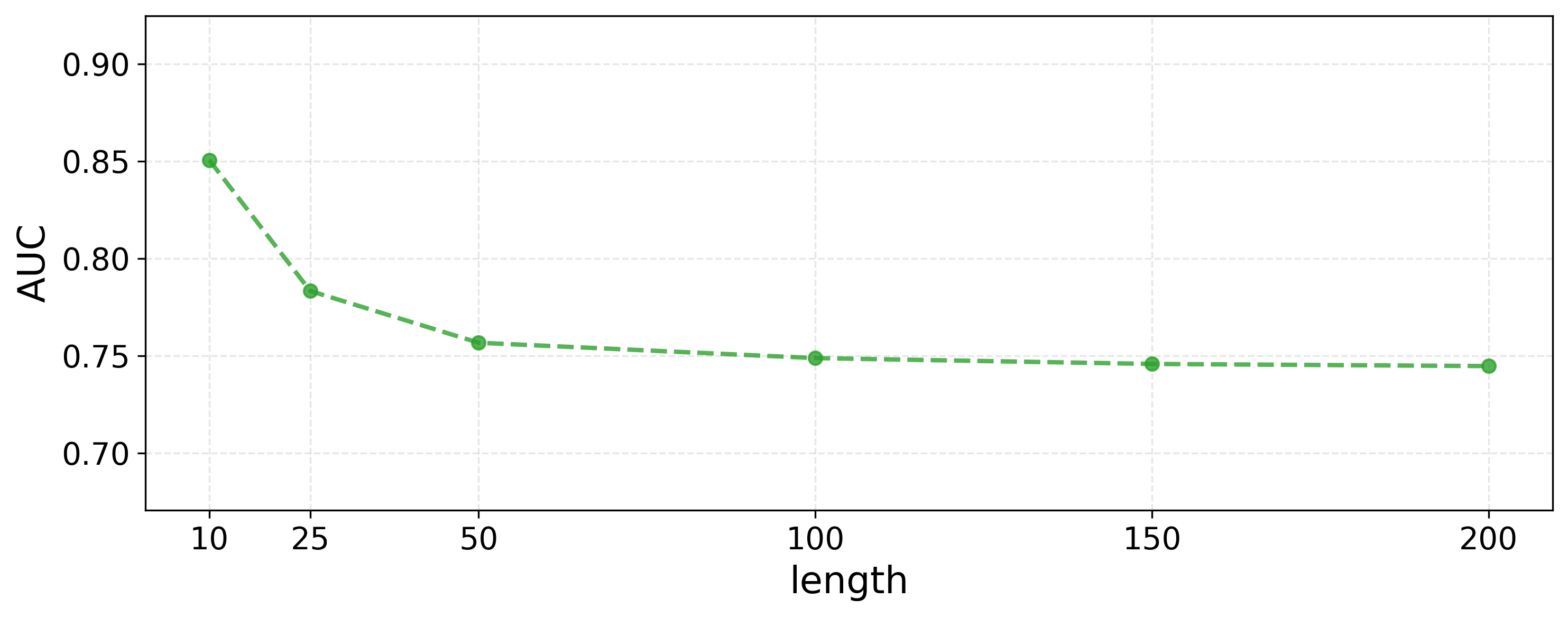}
        \subcaption{The investigation on the length of interactions}
        \label{lossWeightLineFigure-a}
      \end{subfigure}
    }
    \par\bigskip  % 竖直间距，可按需调整
    \scalebox{0.999}{
      % ---------- 子图 b ----------
      \begin{subfigure}{1\linewidth}
        \centering
        \includegraphics[width=\linewidth]{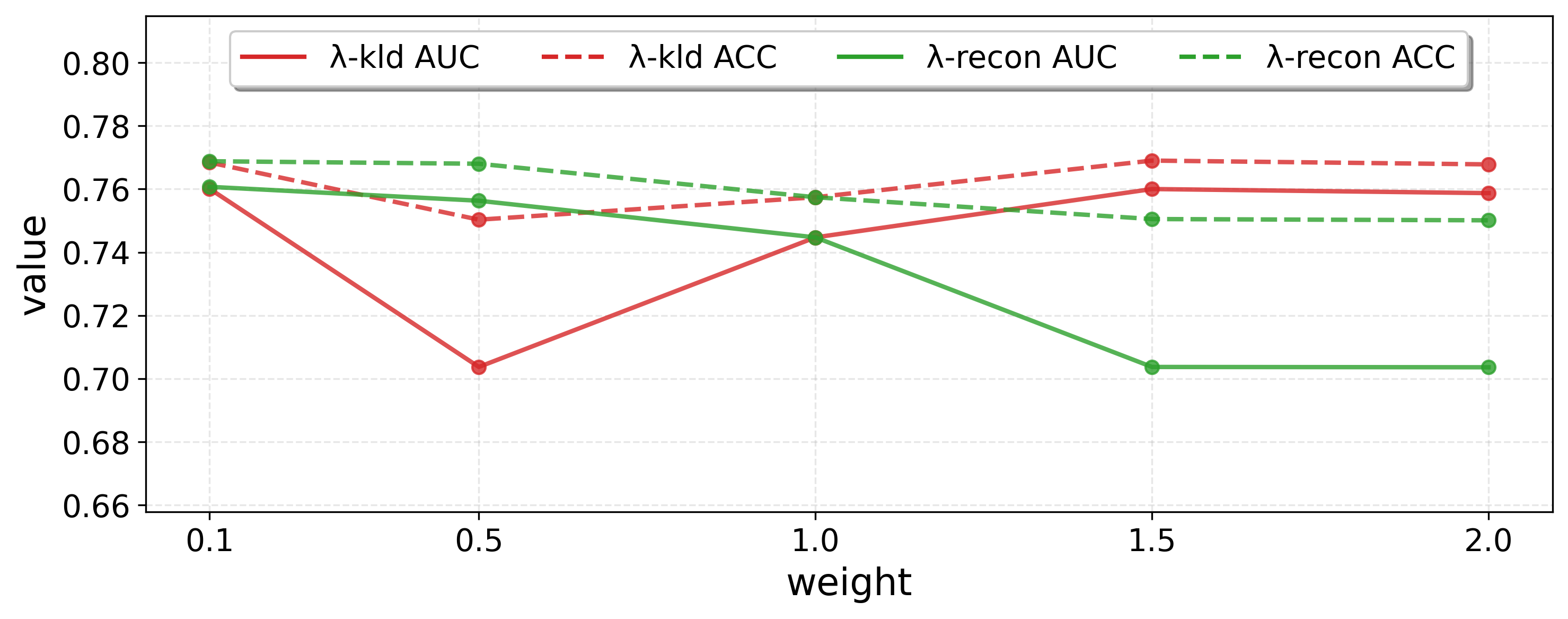}
        \subcaption{The investigation on $\lambda_{kld}$ and $\lambda_{rec}$}
        \label{lossWeightLineFigure-b}
      \end{subfigure}
    }
  \caption{The investigation on hyperparameter}
  \label{lossWeightLineFigure}
\end{figure}

\subsubsection{Sensitivity Analysis of Hyperparameter}
We conducted a parameter sensitivity investigation on the main parameters in memoryKT. First, we explore the influence of sequence length on the ASSIST15 dataset, as shown in the Fig.~\ref{lossWeightLineFigure-a}, memoryKT stops the downward trend at length 50 and tends to stabilize. Therefore, we set the sequence length to 50 and continue exploring the weight settings for $\lambda_{kld}$ and $\lambda_{rec}$. Specifically, we fix the weight of the main task's objective function, the prediction loss, at 1.0. Initially, we set the weight of $\lambda_{kld}$ to 1.0, while the weight of $\lambda_{rec}$ takes values from 0.1, 0.5, 1.0, 1.5, to 2.0 sequentially. Subsequently, we swap the $\lambda_{kld}$ and $\lambda_{rec}$ and repeat this process. As shown in the Fig.~\ref{lossWeightLineFigure-b}, we have the following observations: (1) Although both ACC lines appear very stable overall, a closer look reveals that both lines fluctuate in the same way as the corresponding AUC lines. (2) Generally, the line for $\lambda_{kld}$ is low on the left and high on the right, while the line for $\lambda_{rec}$ is high on the left and low on the right. We believe this is because the temporal information and randomness controlled by $\lambda_{kld}$ should not be too small, while the generator represented by $\lambda_{rec}$ should not impose too much burden on the model. In summary, $\lambda_{kld}$ is suitable for larger values, while $\lambda_{rec}$ is suitable for smaller values.

\begin{figure}[htbp]
  \centering
  \scalebox{0.789}{
      % ---------- 子图 a ----------
      \begin{subfigure}{1\linewidth}  % 宽度占满整行
        \centering
        \includegraphics[width=\linewidth]{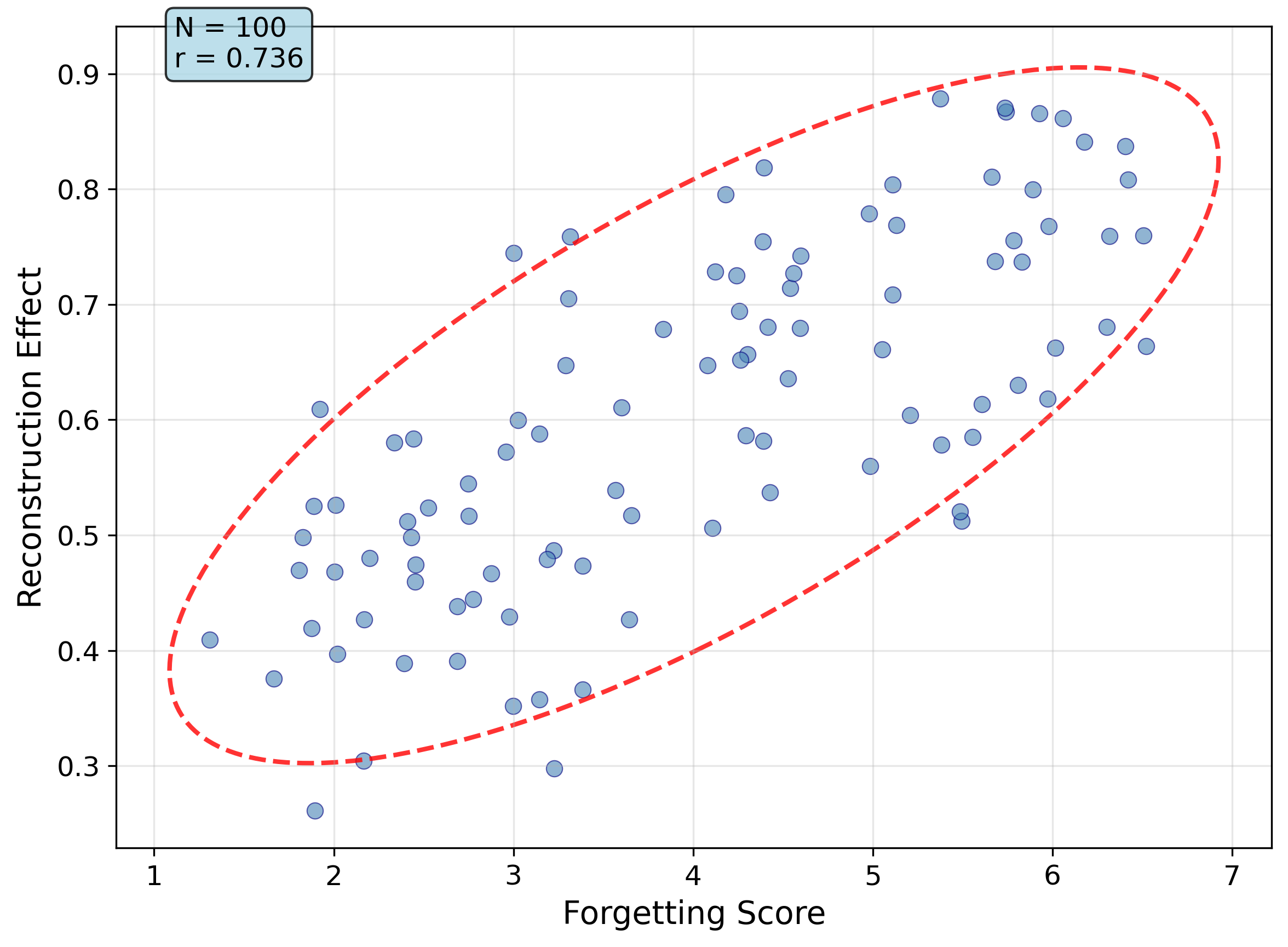}
        \subcaption{reconstruction vs forgetting}
        \label{case study:sub-a}
      \end{subfigure}
      \par\bigskip  % 竖直间距，可按需调整
    }
    \scalebox{0.789}{
    \
      % ---------- 子图 b ----------
      \begin{subfigure}{1\linewidth}
        \centering
        \includegraphics[width=\linewidth]{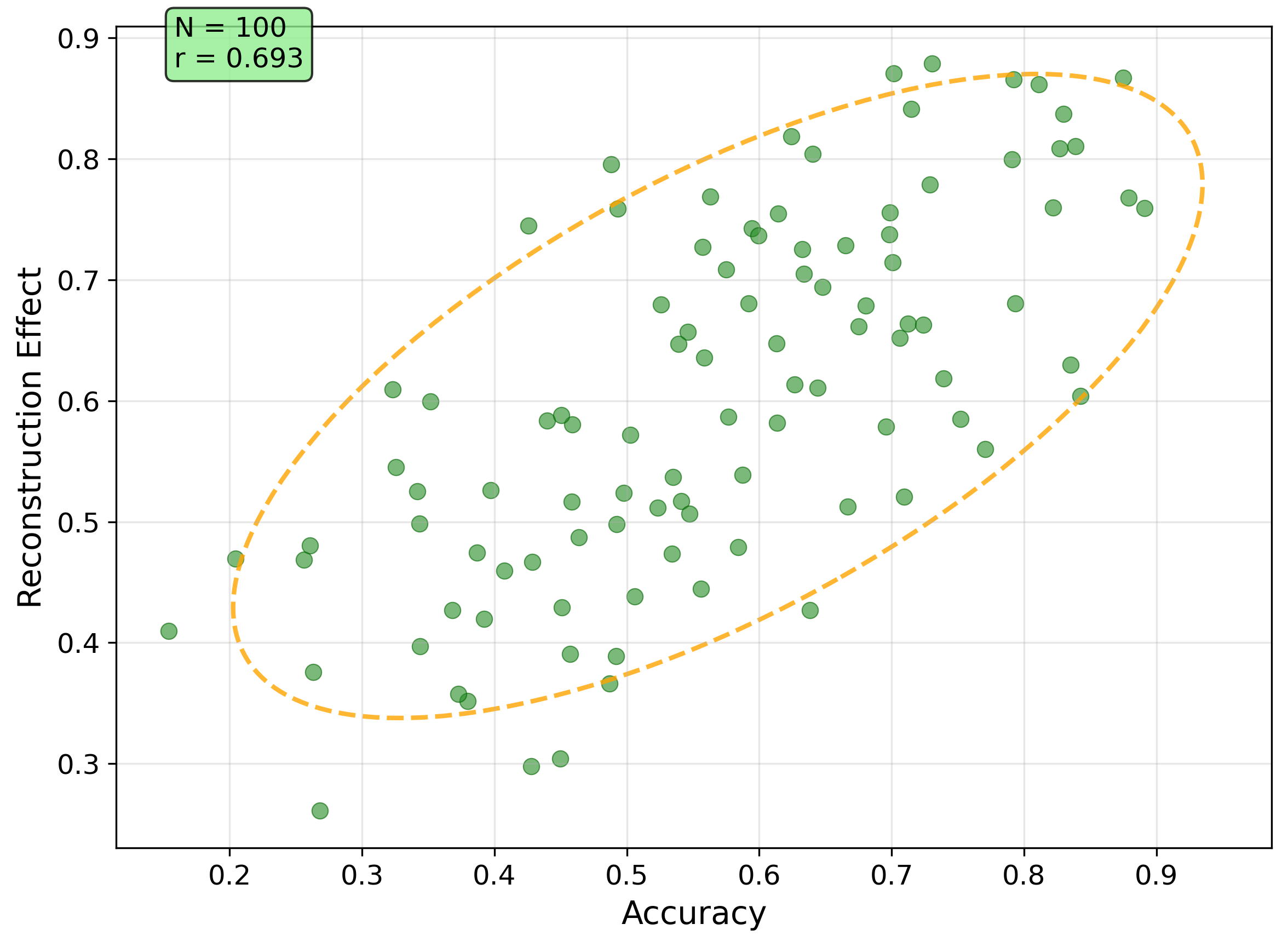}
        \subcaption{reconstruction vs accuracy}
        \label{case study:sub-b}
      \end{subfigure}
    }
  \caption{Study for reconstruction effect and forgetting score}
  \label{case study}
\end{figure}

\subsubsection{Case Study}
In order to explore the model's ability to capture students' memory, we randomly sampled 100 students from the ASISST15 dataset and processed their answer sequences individually. Specifically, we used the trained model to obtain the average interaction reconstruction effect of each student and normalized it so that the larger the processed value, the better the reconstruction effect. In addition, we used the personalized forgetting algorithm to calculate each student's forgetting score, where higher scores indicate better memory, and calculated the correct response rate of each student. As shown in Fig.~\ref{case study}, after obtaining 100 students and their three data points, we plot two scatter plots, where each point represents a student, $N$ is the number of students, and $r$ is the Pearson correlation coefficient. Fig.~\ref{case study:sub-a} shows a strong correlation between the students' interactive reconstruction effect and the forgetting score. Fig.~\ref{case study:sub-b} shows a positive correlation between the students' interaction reconstruction effect and the correct response rate. We believe that students' memory is reflected in two aspects: on the one hand, students' ability to remember knowledge when learning, and on the other hand, the ability to retain memory after a period of time. The consistency of the three data points shows that memoryKT integrates these two aspects to achieve the modeling of student memory. In addition, we found that the value of the reconstruction effect of student interaction before normalization is particularly high, on the one hand, because of the real-time reconstruction of each time step, and on the other hand, it may be that the simple data reconstruction of knowledge tracing is relatively simple, and how to enrich the interactive information remembered by students is worth studying in the future.

\section{Conclusion}
%In this paper, we propose a novel approach to jointly model the complete encoding-storage-retrieval cycle of students’ memory process in learning, which is based on LSTM with the VAE generator and a personal forgetting algorithm that is suitable for the laws of forgetting. 
Based on the idea that memory is an active generation system rather than passive storage, this paper proposes a novel knowledge tracing model that simulates the three-stage memory process of encoding, storage, and retrieval. 
%The model incorporates an LSTM-VAE generator and establishes a personalized forgetting algorithm adapted to the patterns of forgetting.
MemoryKT adopts a VAE variant suitable for sequence prediction to decouple students' personalized memory distribution from the answer history. In addition, the model also adopts an efficient statistical strategy to form a personalized forgetting score, which further improves the personalized modeling ability of the KT model.
%As the first KT method to simultaneously model the three links of memory, memoryKT, this integrative memory-and-forgetting method significantly enhances the model’s performance and interpretability, and also enhances the model’s perception capability for individual differences. 
Extensive experiments on four public datasets demonstrate that memoryKT significantly outperforms state-of-the-art baselines.

%%%%%%%%%%%%%%   Bibliography   %%%%%%%%%%%%%%
% \normalsize
% \bibliography{main}
% \bibliography{reference}

%%%%%%%%%%%%  Supplementary Figures  %%%%%%%%%%%%
% \clearpage

%%%%%%%%%%%% Supplementary Methods %%%%%%%%%%%%

%%%%%%%%%%%%%%%%   End   %%%%%%%%%%%%%%%%
%\end{multicols}  % Method B for two-column formatting (doesn't play well with line numbers), comment out if using method A
\newpage
\bibliography{reference}
\end{document}